\documentclass[letterpaper, 10 pt, journal, twoside]{IEEEtran}
\usepackage{hyperref}

\newif\ifarxiv
\arxivfalse

\newif\ifcaptionmod
\captionmodfalse

\IEEEoverridecommandlockouts                              % This command is only needed if                          % you want to use the \thanks command

\makeatletter
\def\input@path{{.}{Paper/}{Sections/}{Paper/Sections}}
\makeatother

\usepackage{graphics,graphicx} % for pdf, bitmapped graphics files
\graphicspath{{Figures/}}

\usepackage{amsmath} % assumes amsmath package installed
\usepackage{amssymb}  % assumes amsmath package installed
\usepackage{xcolor}

\usepackage{amsmath} % assumes amsmath package installed
\usepackage{amssymb}  % assumes amsmath package installed
\usepackage{xcolor}
\usepackage{algorithm}
\usepackage{algpseudocode}
\usepackage{cite}
\usepackage{dblfloatfix}
\usepackage{mathtools} % include prescript
\usepackage{dutchcal} % caligraphy
\usepackage{caption}
\usepackage{subcaption}
\usepackage{multirow}
\usepackage{lipsum}

\renewcommand{\xi}{\x^{[i]}}
\renewcommand{\H}{\mathbf{H}}
\renewcommand{\v}{\mathbf{v}}
\renewcommand{\S}{\mathbf{S}}

\renewcommand{\u}{\mathbf{u}}
\renewcommand{\P}{\mathbf{P}}
\renewcommand{\S}{\mathbf{S}}
\renewcommand{\u}{\mathbf{u}}
\renewcommand{\c}{\mathbf{c}}
\renewcommand{\a}{\mathbf{a}}

\newcommand{\norm}[1]{\left\lVert#1\right\lVert}

\newcommand{\x}{\mathbf{x}}

\newcommand{\f}{\mathbf{f}}
\newcommand{\fx}{\mathbf{f}_{\mathbf{x}}}
\newcommand{\fu}{\mathbf{f}_{\mathbf{u}}}
\newcommand{\fxx}{\mathbf{f}_{\mathbf{_{xx}}}}
\newcommand{\fuu}{\mathbf{f}_{\mathbf{{uu}}}}
\newcommand{\fux}{\mathbf{f}_{\mathbf{{ux}}}}

\newcommand{\inv}[1]{{#1}^{-1}}
\newcommand{\Qx}{\mathbf{Q}_{\x}} 
\newcommand{\Qu}{\mathbf{Q}_{\u}}
\newcommand{\Qxx}{\mathbf{Q}_{\x\x}}
\newcommand{\Qux}{\mathbf{Q}_{\u\x}}
\newcommand{\Quu}{\mathbf{Q}_{\u\u}}
\newcommand{\Lx}{\mathbf{L}_{\x}}
\newcommand{\Lu}{\mathbf{L}_{\u}}
\newcommand{\Lxx}{\mathbf{L}_{\x\x}}
\newcommand{\Luu}{\mathbf{L}_{\u\u}}
\newcommand{\Lux}{\mathbf{L}_{\u\x}}
\newcommand{\qs}{\mathbf{q}}
\newcommand{\U}{\mathbf{U}}
\newcommand{\g}{\mathbf{g}}
\newcommand{\bcalU}{\mathbcal{U}}
\newcommand{\blambda}{\boldsymbol{\lambda}}
\newcommand{\bsigma}{\boldsymbol{\sigma}}
\newcommand{\bdelta}{\boldsymbol{\delta}}
\newcommand{\calL}{\mathcal{L}}
\newcommand{\Si}{\mathbf{S}^{[i]}}
\newcommand{\bsi}{\mathbf{s}^{[i]}}
\newcommand{\si}{s^{[i]}}
\newcommand{\s}{\mathbf{s}}
\newcommand{\bkappa}{\boldsymbol{\kappa}}
\newcommand{\R}{\mathbf{R}}
\newcommand{\p}{\mathbf{p}}
\newcommand{\T}{\mathbf{T}}

\newcommand{\J}{\mathbf{J}}
\newcommand{\C}{\mathbf{C}}
\newcommand{\taub}{\boldsymbol{\tau}}

\newcommand{\I}{\mathbf{I}}
\newcommand{\h}{\mathbf{h}}
\newcommand{\ui}{\u^{[i]}}
\newcommand{\bomega}{\boldsymbol{\omega}}
\newcommand{\Icom}{{}^b\overline{\mathbf{I}}}
\newcommand{\quat}{\boldsymbol{q}}
\newcommand{\CoM}{\text{CoM}}

\newcommand{\Real}{\mathbb{R}}
\usepackage{qting}

\title{\LARGE \bf
Model Hierarchy Predictive Control of Robotic Systems
}

\author{He Li, Robert J. Frei, Patrick M. Wensing\\ \vspace{-1ex} % <-this % stops a space
\thanks{This work was supported by NSF Grant CMMI-1835186 and ONR Award N0001420WX01278.}
\thanks{Authors are with Department of Aerospace and Mechanical Engineering,
        University of Notre Dame, Notre Dame, IN 46556 USA
        ({\tt\small hli25@nd.edu, rfrei@nd.edu pwensing@nd.edu})}%
}

\begin{document}

\maketitle
% Comment below for final MS
% \thispagestyle{plain}
% \pagestyle{plain}

%%%%%%%%%%%%%%%%%%%%%%%%%%%%%%%%%%%%%%%%%%%%%%%%%%%%%%%%%%%%%%%%%%%%%%%%%%%%%%%%
\begin{abstract}
This paper presents a new predictive control architecture for high-dimensional robotic systems. As opposed to a conventional Model Predictive Control (MPC) approach to locomotion that formulates a hierarchical sequence of optimization problems, the proposed work formulates a single optimization problem posed over a hierarchy of models, and is thus named Model Hierarchy Predictive Control (MHPC). MHPC is formulated as a multi-phase receding-horizon Trajectory Optimization (TO) problem, and can be implemented using any general multi-phase TO solver. MHPC is benchmarked in simulation on a quadruped, a biped, and a quadrotor, demonstrating control performance on par or exceeding whole-body MPC while maintaining a lower computational cost in each case. A preliminary gap jumping experiment is conducted on the MIT Mini Cheetah with the control policy generated offline, demonstrating the physical validity of the generated trajectories and motivating online MHPC in future work.
\end{abstract}


%%%%%%%%%%%%%%%%%%%%%%%%%%%%%%%%%%%%%%%%%%%%%%%%%%%%%%%%%%%%%%%%%%%%%%%%%%%%%%%%

\section{INTRODUCTION}
\label{sec:Intro}

\IEEEPARstart{W}{hether} for agriculture, construction, or disaster response, the mobility afforded by legs offers promise for future robots that can go where we go. Despite rapid progress in the past decade toward this vision, the operational envelope of existing legged robots remains limited. Time-consuming tuning in the lab is usually necessary for every new behavior and every new robot. While this approach may be acceptable for current robots to operate in known scenarios, future robots must master the ability to generate stable motion on the fly to succeed in navigating novel real-world environments. 

Model Predictive Control (MPC) is a powerful tool in optimal control and has been gaining popularity in the legged robot community over the past decade \cite{tassa2012synthesis,wieber2006trajectory,di2018dynamic, neunert2018whole}. Compared to conventional optimal control methods such as the Linear Quadratic Regulator (LQR), MPC considers system dynamics and other constraints, e.g., control and path limits, while repeatedly solving a finite-horizon trajectory optimization (TO) problem from the current state. The first control signal is executed and the next state is measured. The TO is then re-solved with the new current state. The performance of MPC depends on the dynamics model used, horizon length, and update frequency. This work proposes a new MPC configuration by using hierarchical model structures over the planning horizon to accelerate the solution of these problems.

Current MPC for legged robots mainly takes one of two approaches, simple-model MPC or whole-body MPC. Simple-model MPC formulates a sequence of optimization problems. This is done by generating a long-term low-dimensional plan based on a simple model, then solving a quadratic optimization (QP) problem at every instant for the full model to track the plan. Commonly used simple models include the Linear Inverted Pendulum Model (LIP) \cite{wieber2006trajectory,bellicoso2018dynamic}, Spring-Loaded Inverted Pendulum Model (SLIP) \cite{wensing2013high}, and Single-Rigid-Body Model \cite{bledt2017policy,winkler2018gait,di2018dynamic}. Centroidal dynamics models have also been used that consider the linear and angular momentum of the system as a whole \cite{orin2013centroidal,dai2014whole, wensing2016improved}. Simple-model MPC has fast computation, and it is often easier to find a good initial guess compared to whole-body MPC. 
However, there is a complex interplay between short-term constraints posed within the QP and other long-term constraints, which may lead to time-consuming tuning. Additional details beyond the simple model (e.g., swing leg motions) must be separately designed, which adds complexity. Further, the operational envelope of the resulting motions is limited. For example, stair climbing could not be directly generated with a LIP model since it normally neglects all kinematics constraints and assumes constant height and zero angular momentum. 

\begin{figure}[!t]
    \centering
    \includegraphics[width = 0.75\linewidth]{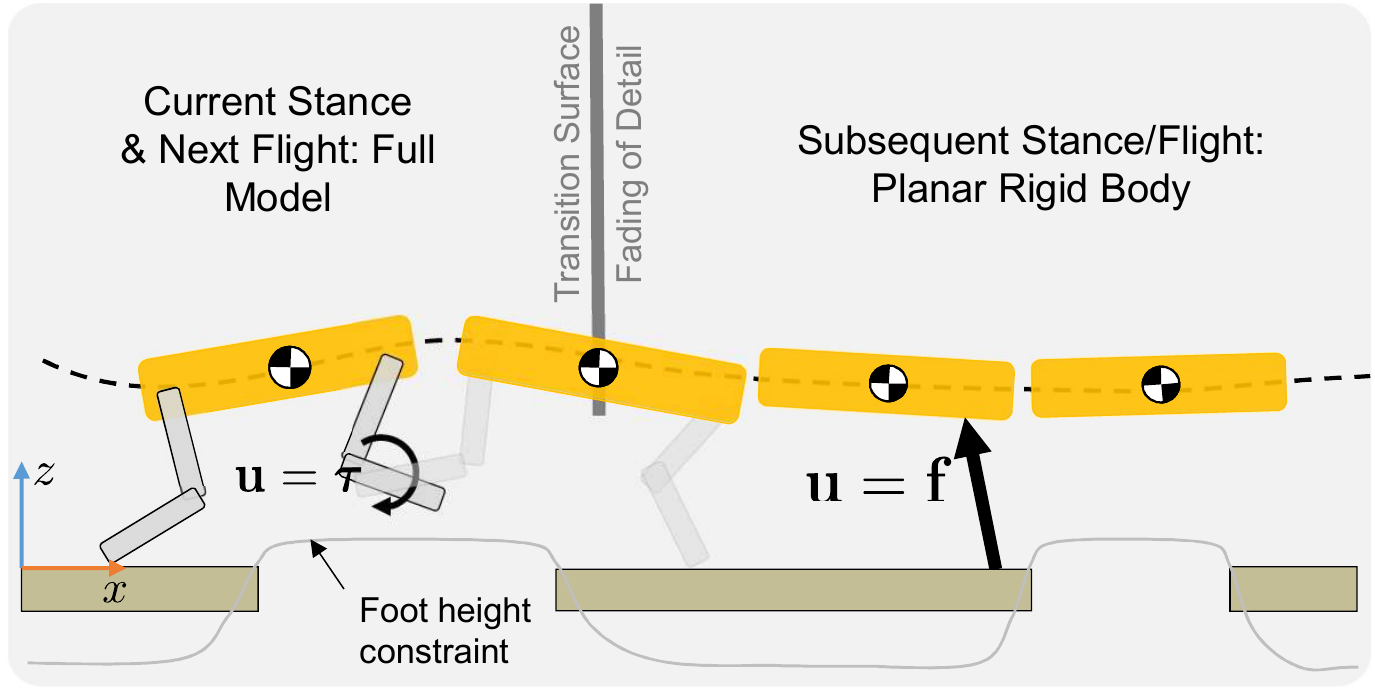}
    \caption{MHPC for a quadruped navigating gaps. MHPC uses multiple models over the planning horizon to capture performance benefits of whole-body MPC with the computational benefits of simple-model planning.}
    \label{fig:MHPC_illusrative}
    \vspace{-13px}
\end{figure}

By comparison, whole-body MPC can generate more complex behaviors by considering the full range of possible motions. Whereas whole-body QPs only consider control effects instantaneously, whole-body MPC finds actuation torques over a time interval that ensures long-term balance, enabling recovery from larger disturbances. Despite these benefits, the \textit{curse of dimensionality} from the high-dimensional dynamics of legged robots has prevented it from being popular, and the underlying TO problem can be highly non-convex, requiring a good initial guess. Recent results (e.g., \cite{tassa2012synthesis}) using Differential Dynamic Programming (DDP) \cite{mayne1966second} have shown great promise for whole-body MPC. Many DDP advances have been subsequently proposed, demonstrating real-time performance in simulation \cite{budhiraja2018differential,mastalli2020crocoddyl} and on hardware \cite{koenemann2015whole, neunert2018whole}. Though some of these works solve a whole-body TO problem for a quadruped, computational requirements are significantly greater than for simple-model MPC. As a result, implementations of whole-body MPC require shorter-sighted planning horizons than with simple-model strategies due to these increased computational demands.

\subsection{Contribution}
\label{subsec:Intro_contribution}
The major contribution of this research is to propose and evaluate Model Hierarchy Predictive Control (MHPC), a new method that carries out planning within MPC based on a hierarchy of models over the planning horizon. This contribution explores the formulation of the MPC problem, and is complementary to many excellent recent advances in numerical solution algorithms (e.g., \cite{mastalli2020crocoddyl,giftthaler2018control}) for optimal control. Overall, MHPC is shown to unify the computational and performance benefits of simple-model MPC and whole-body MPC. Figure~\ref{fig:MHPC_illusrative} conceptually illustrates the main idea for a planar quadruped that needs to account for gaps when planning its movement. In the current stance and next flight, the robot coordinates its legs and body to avoid the gap. Since the subsequent stance and flight are far away, it roughly determines a body motion so that the footsteps can avoid the gap, but it does not propose a concrete plan, i.e., ignores leg details, until the gap comes close. In this way, the quadruped focuses on its near-term balance while having a rough plan in mind for the long term. Building toward this vision, this work first considers a rigorous simulation evaluation for MHPC. As a step toward online MHPC, hardware experiments are then shown for jumping over a gap using the control policies optimized by MHPC in simulation.

%This work makes a first step toward this goal by studying the underlying mechanism, i.e., carrying out MPC with planning over a hierarchy of models.

The rest of this paper is structured as follows. Section~\ref{sec:MHPC} discusses the main idea of MHPC and formulates the MHPC problem mathematically. An efficient DDP-based solver is introduced for MHPC as well. Section~\ref{sec:example_sys} presents a hierarchy of models for three example systems. The performance of MHPC is then benchmarked in simulation in Section~\ref{sec:Sim}. In Section~\ref{sec:Exp}, we run MHPC offline and execute the control policy in a dynamics simulator and on hardware. Section~\ref{sec:Conclusion} concludes the paper and discusses future work.

\section{Model Hierarchy Predictive Control}
\label{sec:MHPC}
%\subsection{Main Idea}
%\label{subsec:MHPC_main_idea}
\begin{figure}[b]
    \center
    \includegraphics[width = 0.95 \linewidth]{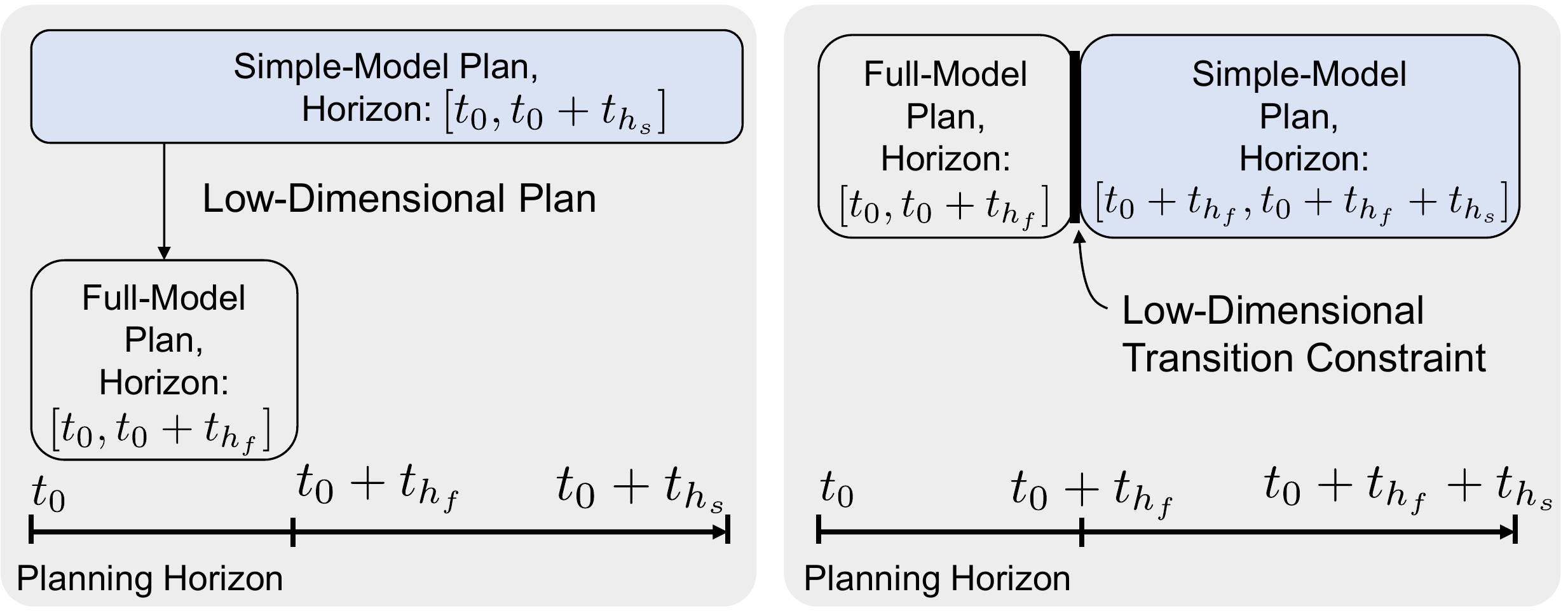}
    \caption{Novelty of MHPC compared to conventional simple-model MPC. Left: Simple-model MPC generates the simple-model plan and the full-model plan via a hierarchical sequence of optimizations. Right: MHPC generates the two plans simultaneously via a single optimization. }
    \label{fig:MHPC_diagram}
\end{figure}

In this section, we further describe the main idea of MHPC. Compared to simple-model MPC, which formulates a hierarchical sequence of optimization problems, MHPC constructs a single optimization problem posed over a hierarchy of models. Main differences are conceptually shown in Fig.~\ref{fig:MHPC_diagram}.  MHPC plans with full-model dynamics in the near term and with a simple model in the long term. The consistency between the two is enforced by a low-dimensional transition constraint. Due to the reduced dimensionality of the simple model, this transition typically incorporates a projection map. This instantaneous transition constraint is in contrast to methods that impose consensus between models over the entire planning horizon (e.g., \cite{budhiraja2019dynamics}). Figure~\ref{fig:MHPC_diagram} incorporates only two models, but, in general, MHPC can use multiple models and regards the higher-order model as a `full model' and lower-order models as `simple models'.

At the core of MHPC is an abstraction schedule specifying the planning horizons for each model. In Fig.~\ref{fig:MHPC_diagram}, $t_{h_f}$ and $t_{h_s}$ respectively represent planning horizon durations for a full model and a simple model. The abstraction schedule affects MHPC performance in terms of disturbance rejection and computational cost. Planning based exclusively on the full model and exclusively on a simple model represent two extremes of how to configure MHPC. In this work, we discuss these effects for several example systems in Section~\ref{sec:example_sys}.

\subsection{Problem Formulation}
\label{subsec:problem formulation}
\vspace{-3px}
\label{subsec:MHPC_problem}
MHPC can be formulated as a multi-phase receding-horizon TO problem, where a phase indicates a period of time during which the dynamics is unchanged. A system with multi-phase dynamics can be modeled as follows:
\begin{subequations}\label{eq_multiplase_system}
\begin{align}
    \xi_{k+1} &= \f_i(\xi_k,\ui_k),\label{subeq_dyn}\\ 
     \x^{[i+1]}_0 &= \P_{i}(\xi_{N_{i}}),\label{subeq_reset}\\
     g_{i}(\xi_{N_{i}}) &= 0, \label{subeq_switch}
\end{align}
\end{subequations}
where $\x$ and $\u$ are system state and control vectors, $i$ denotes the phase index, $k\in [0,N_i-1]$ the time index, $\f$ gives the dynamics evolution, $\P$ represents the phase transition map, and $g$ represents a constraint on the terminal state. The state and control vectors may have a different dimension in each phase. Note that $\f$ is in discrete time and is obtained via a numerical integration scheme, with forward Euler used in this work. Other integration schemes can be used without loss of generality to the remaining development. The equation~\eqref{subeq_switch} specifies a terminal constraint for the $i$-th phase. 

We alert the readers that the formulation~\eqref{eq_multiplase_system} is a generalization of other hybrid systems models commonly used in legged locomotion work (e.g., \cite{westervelt2003hybrid}) where the transition map normally accounts for impacts. In~\eqref{eq_multiplase_system}, $\f$ may specify the dynamics for a phase of the full model (e.g., stance dynamics or flight dynamics of a full quadruped) or of a simple model. The transition map $\P$ may thus describe a reset map at impact, as well as a transition that maps the state of the full model to the state of a simpler model, as captured by the transition constraint in Fig.~\ref{fig:MHPC_diagram}. For example, when working with the SLIP model, the transition map could account for the impact in the full model and then extract the CoM state post-impact as the initial condition for the SLIP.

For each phase $i$ of~\eqref{eq_multiplase_system}, we define a cost function
\begin{equation}
    J^{[i]}(\x^{[i]}_0, \U^{[i]}) = \sum_{k=0}^{N_i - 1} \ell_i(\xi_k, \ui_k) + \phi_i(\xi_{N_i}),
\end{equation}
where $\ell_i$, $\phi_i$ and $N_i$ represent the running cost, the terminal cost, and the length of horizon, respectively, associated with phase $i$, and $\U^{[i]} = [\u^{[i]}_0,\cdots,\u^{[i]}_{N_i-1}]$. For a system~\eqref{eq_multiplase_system} with $n$ phases, the TO problem can be formulated as follows 
\begin{subequations}\label{eq_hybridTO}
\begin{IEEEeqnarray}{cll}
\IEEEeqnarraymulticol{2}{l}{\min_{\bcalU} \ \ \sum_{i=1}^n J^{[i]}(\x^{[i]}_0,\U^{[i]})}\\
    \text{subject~to} \ \ & \eqref{subeq_dyn}, \eqref{subeq_reset},\\ 
     & \eqref{subeq_switch},\\
     &\h_i(\xi_k,\ui_k)\geq \mathbf{0}, \label{subeq_ineq}
     \end{IEEEeqnarray}
     \end{subequations}
where $\bcalU = [\U^{[1]},\cdots,\U^{[n]}]$ and $\h$ describes inequality constraints such as torque limits and friction constraints. Solving~\eqref{eq_hybridTO} with receding horizons then gives rise to MHPC. Extending the horizon of a conventional finite-horizon Optimal Control Problem (OCP) helps maintain long-term stability at the expense of additional computation. For example, the whole-body QP designed in \cite{Sherikov14} adds long-term balance constraints for a humanoid robot such that its state remains viable. An alternate strategy is to embed future costs into a terminal cost. An ideal terminal cost would be the value function of an infinite-horizon OCP, which, however, does not have an analytical solution in general and needs to be approximated. In this work, by planning over the long term with a simple model, MHPC provides a proxy of the value function for the full model to improve its long-term stability.

\subsection{Hybrid Systems Solver for MHPC}
\label{subsec:MHPC_solver}
We employ an efficient algorithm, Hybrid Systems Differential Dynamic Programming (HSDDP), developed in \cite{li2020hybrid}, to solve the multi-phase optimization problem~\eqref{eq_hybridTO}. Other general multi-phase TO solvers could also be used. HSDDP attacks~\eqref{eq_hybridTO} by converting it into an unconstrained optimization problem using Augmented Lagrangian (AL) \cite{lantoine2012hybrid,Howell19} and Reduced Barrier (ReB) methods. A Lagrangian-like function for phase $i$ is constructed as follows
\begin{multline}\label{eq_mode_Lagrangian}
    \calL^{[i]}(\x^{[i]}_0,\U^{[i]},\lambda_i,\sigma_i,\delta_i) = \\ \sum_{k=0}^{N_i - 1} \underbrace{\ell_i(\xi_k, \ui_k) + B_{\delta_i}\big(\h_i(\xi_k,\ui_k) \big)}_{L_i(\xi_k, \ui_k)} \\ 
    +\underbrace{\phi_i(\xi_{N_i}) + \left(\frac{\sigma_i}{2}\right)^2 g_i^2(\xi_{N_i}) + \lambda_i g_i(\xi_{N_i})}_{\Phi_i(\xi_{N_i})},
\end{multline}
where $\lambda_i$ and $\sigma_i$ are the Lagrange multiplier and penalty coefficient, respectively, and $B_{\delta_i}$ is an element-wise Reduced Barrier function \cite{hauser2006barrier,li2020hybrid} with relaxation parameter $\delta_i$. The Lagrangian-like function for problem~\eqref{eq_hybridTO} is then
\begin{equation}\label{eq_Lagrangian}
    \calL(\bcalU,\blambda, \bsigma, \bdelta) =  \sum_{i=1}^n \calL^{[i]}(\x^{[i]}_0,\U^{[i]},\lambda_i,\sigma_i,\delta_i),
\end{equation}
where $\blambda = [\lambda_1,\cdots,\lambda_n]$, $\bsigma$ and $\bdelta$ are similarly defined. 

The reformulated unconstrained problem is then:
\begin{subequations}\label{eq_hybridTO_unconstr}
\begin{IEEEeqnarray}{cll}
\IEEEeqnarraymulticol{2}{l}{\min_{\bcalU} \ \ \calL(\bcalU,\blambda, \bsigma, \bdelta)}\\
    \text{subject~to} \ \ & \eqref{subeq_dyn}, \eqref{subeq_reset}.
\end{IEEEeqnarray}
\end{subequations}
HSDDP solves~\eqref{eq_hybridTO_unconstr} with fixed $\blambda, \bsigma, \bdelta$ using DDP, and employs an outer loop to iteratively adjust their values until all constraints are satisfied. The update equations are as follows
\begin{equation}
    \bsigma \leftarrow \beta_{\sigma}\bsigma, \ \ 
    \blambda \leftarrow \blambda + \bsigma \circ\g, \ \   
    \bdelta \leftarrow \beta_{\delta}\bdelta, \label{eq_update}
\end{equation}
where $\beta_{\sigma}>1$ and $0<\beta_{\delta}<1$ are update parameters, the vector $\g\in\Real^n$ concatenates $g_{i}(\xi_{N_{i}})$ $\forall i=1,\cdots,n$, and the operator $\circ$ denotes element-wise product. We note that the convergence of this strategy to a feasible point is not strictly guaranteed, however, it is empirically found to be effective. Care must be taken when solving~\eqref{eq_hybridTO_unconstr} with DDP due to the discontinuous jump (for hybrid systems) or state projection (for model transition) caused by~\eqref{subeq_reset}. This is addressed in HSDDP by employing an impact-aware step~\cite{li2020hybrid}. 

We denote $V(\xi_k)$ the optimal cost-to-go evaluated at $\xi_k$. The variation of $V(\xi_k)$ along a nominal trajectory under perturbation $\delta\xi_k$ is approximated to the second order as follows
\begin{equation}\label{eq_value_func_approx}
    \delta V(\delta\xi_k) \approx \frac{1}{2}(\delta\xi_k)^\top\Si_k\delta\xi_k + (\bsi_k)^\top\delta\xi_k + \si_k,
\end{equation}
where $\Si_k$, $\bsi_k$, and $\si_k$ respectively, represent the Hessian, gradient, and scalar terms. HSDDP recursively computes these terms for all $k$ and $i$ by performing a backward sweep. For $0\leq k<N_i$ in phase $i$, the recursive equations are \cite{tassa2012synthesis}
\begin{subequations}\label{eq_value_update_smooth}
\begin{align}
    s &= s'  - \frac{1}{2}\Qu^T\Quu^{-1}\Qu, \label{eq_scalar}\\
    \s &= \Qx - \Qux^T\Quu^{-1}\Qu, \label{eq_gradient}\\
    \S &= \Qxx - \Qux^T\Quu^{-1}\Qux \label{eq_hessain},
\end{align}
\end{subequations}
in which
\begin{subequations}\label{eq_Qs}
\begin{align}
    \Qx &= \Lx + \fx^\top\s',\\
    \Qu &= \Lu + \fu^\top\s',\\
    \Qxx &= \Lxx + \fx^\top\S'\fx + \s' \cdot  \fxx,\\
    \Quu &= \Luu + \fu^\top\S'\fu + \s' \cdot  \fuu,\\
    \Qux &= \Lux + \fu^\top\S'\fx + \s' \cdot \fux,
\end{align}
\end{subequations}
where $\mathbf{L}$ indicates derivatives of $\calL$, the subscripts indicate partial derivatives and the prime indicates the next time step. Note that $\fxx$, $\fuu$ and $\fux$ are tensors. The notation `$\cdot$' denotes vector-tensor multiplication. At $k=N_i$ of phase $i$, the update equations are
\begin{subequations}\label{eq_value_update_jump}
\begin{align}
    s &= s',\\
    \s &= \boldsymbol{\Phi}_{\x} + \P_{\x}^\top\s',\\
    \S & = \boldsymbol{\Phi}_{\x\x} + \P_{\x}^\top\S' \P_{\x}.
\end{align}
\end{subequations}
Note that in ~\eqref{eq_value_update_jump}, the prime indicates the next step at $k=0$ of phase $i+1$. The value function approximation thus obtained is used to construct the update policy 
\begin{equation}\label{eq_optdu}
     \delta\u^* = -\Quu^{-1}(\Qu + \Qux\delta\x) \equiv \bkappa + \mathbf{K}\delta\x.
\end{equation}
where $\bkappa$ is the search direction (feed-forward correction) and $\mathbf{K}$ is a feedback gain.
A line search method and regularization are typically performed with~\eqref{eq_optdu} to ensure cost reduction \cite{tassa2012synthesis}. Pseudocode for HSDDP is shown in Alg.~\ref{alg_HSDDP}, and the reader is referred to \cite{li2020hybrid} for a detailed description.

\begin{algorithm}[t]
    \caption{HSDDP Algorithm}
    \label{alg_HSDDP}
    \begin{algorithmic}[1] 
        \State Initialize $\blambda,\bsigma, \bdelta$.
        \While {$\norm{\g}_2>$ tol.}
        \State Minimize $\calL(\bcalU,\blambda,\bsigma, \bdelta)$ using DDP (using Eq.~\eqref{eq_value_update_jump} at each mode transition in the backward sweep).
        \State Compute $g_{c_i}(\x^-_{N_i})$.
        \State Update $\blambda,\bsigma, \bdelta$ using Eq.~\eqref{eq_update}
        \EndWhile
    \end{algorithmic}
\end{algorithm}

\section{Hierarchy of Models and Abstraction Schedules for Example Systems}
\label{sec:example_sys}
This section presents model hierarchies for MHPC of a planar quadruped, a five-link biped, and a quadrotor. 

\subsection{Planar quadruped and biped}\label{subsec_quadruped}
\begin{figure}[b]
    \centering
    \includegraphics[width =0.95\linewidth]{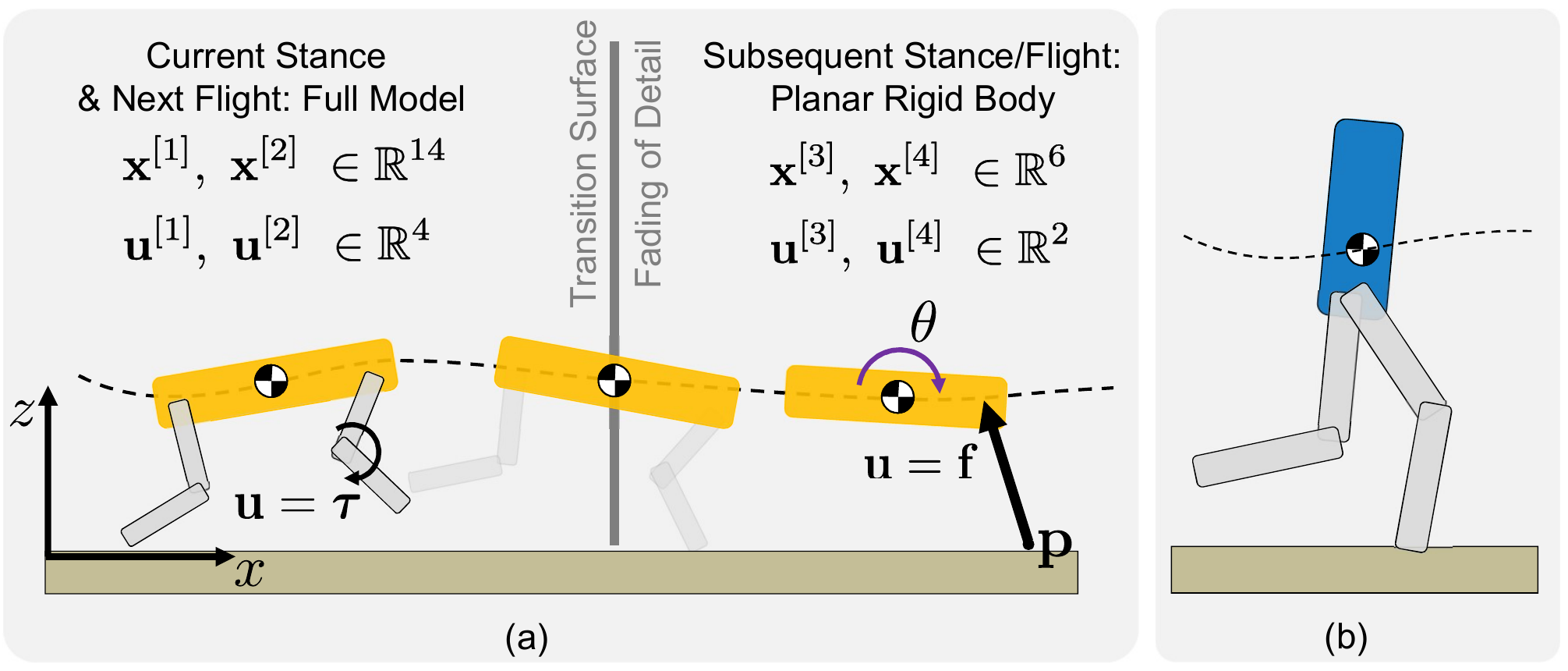}
    \caption{Hierarchy of models for five-link robots with MHPC. (a) detailed abstraction schedule for planar quadruped bounding (b) the same abstraction is applied for biped running.}
    \label{fig:hierarcy_legged}
\end{figure}

The dynamics of a planar quadruped are hybrid since it makes and breaks contact with ground. As a result, the full model of a quadruped could be described by system~\eqref{eq_multiplase_system} whose dynamics in each phase are obtained as \cite{budhiraja2018differential, li2020hybrid}
\begin{equation}\label{eq_fullmodel_smooth}
        \begin{bmatrix}
                \H & -\J_{c}^\top \\
                -\J_{c} & \mathbf{0}
        \end{bmatrix}
            \begin{bmatrix}
                \Ddot{\qs} \\
                \blambda_{c}
            \end{bmatrix}
            =   \begin{bmatrix}
                    \S^\top\taub - \mathbf{C}\Dot{\qs} - \taub_g\\
                    \Dot{\J}_{c}\Dot{\qs}
            \end{bmatrix},
\end{equation}
where $\qs = [\c^\top, \theta, \qs^\top_{\text{joint}}]^\top$ is the generalized coordinate, $\c\in\Real^2$, $\theta \in \Real$, and $\qs_{\text{joint}}\in\Real^4$ denote the trunk center of mass (CoM), trunk orientation, and joint angles, respectively, and $\H$, $\C \Dot{\qs}$, $\taub_g$, $\S$, and $\taub\in\Real^4$ denote the inertia matrix, Coriolis force, gravity force, selection matrix, and actuation torque, respectively. $\J_{c}$ and $\blambda_{c}$ represent the contact Jacobian and contact force, respectively, and their expressions depend on which foot is in contact with the ground. During a flight mode, the matrix on the left side of~\eqref{eq_fullmodel_smooth} degenerates to the inertia matrix $\H$. Denoting $\x_f = [\qs^\top, \Dot{\qs}^\top]^\top$ as the full-model state, the state-space representation of~\eqref{eq_fullmodel_smooth} can then readily be obtained into the form~\eqref{eq_multiplase_system}. The reset map at touchdown is modeled considering impact dynamics as follows
\begin{equation}\label{eq_fullmode_impact}
    \P_{\text{TD}}(\x_f^{-}) = 
    \begin{bmatrix}
        \I & \mathbf{0}\\
        \mathbf{0} & \I-\inv{\H}\J_{c}^\top(\J_{c}\inv{\H}\J_c^\top)^{-1}\J_c
    \end{bmatrix}
    \x_f^-,
\end{equation}
where the superscript `-' indicates the moment immediately before contact. The reset map at liftoff is smooth and simply is $\P_{\text{LO}}(\x_f^-)  = \x_f^-$, where `-' indicates the moment immediately before breaking contact.

For a quadruped robot, the weight of its legs is typically a small fraction of its total weight. As result, a common simple model ignores the legs and considers only trunk dynamics: 
\begin{equation}\label{eq_trunk_model}
    \begin{aligned}
    \Ddot{\c} &= \sum_i \frac{\f_i}{m} -\g \\
    I \Ddot{\theta} &= \sum_i (\p_i - \c) \times \f_i ,
    \end{aligned}
\end{equation}
where $\c$, $\theta$, and $I$ represent the position of the CoM, the trunk orientation, and the body inertia, respectively. $\f$, $\g$ and $\p$ denote the ground reaction force (GRF), gravitational acceleration, and the foot location, respectively, and $i$ indicates the contact foot index. We denote the state of this trunk model as $\x_s = [\c,\theta, \Dot{\c}, \Dot{\theta}]$. A projection map relating the full and simple models is then defined such that $\x_s = \T\x_f$ where 
\begin{equation}\label{eq_contraction}
    \T = 
    \begin{bmatrix} 
    \I^{3} & \mathbf{0}^{3\times 4} & \mathbf{0}^{3\times3} & \mathbf{0}^{3\times 4} \\
    \mathbf{0}^{3\times 3} & \mathbf{0}^{3\times 4} & \I^{3} & \mathbf{0}^{3\times 4}
    \end{bmatrix}
\end{equation}
and $\I^3\in\Real^{3\times3}$ is an identity matrix.

We now define the low-dimensional transition constraint as in Fig.~\ref{fig:MHPC_diagram}. Take quadruped bounding as an example, which periodically executes four gait modes of motion: a back stance mode, a flight mode, a front stance mode, and another flight mode. Figure~\ref{fig:hierarcy_legged} illustrates an abstraction schedule used for quadruped bounding that assigns the first two modes for the full model~\eqref{eq_fullmodel_smooth}, \eqref{eq_fullmode_impact}, and the subsequent two modes for the trunk model~\eqref{eq_trunk_model}. At the instant of touchdown shown in Fig.~\ref{fig:hierarcy_legged}(a), a transition from full model to trunk model takes place. The low-dimensional transition thus considers the impact discontinuity~\eqref{eq_fullmode_impact} caused by touchdown, as well as the projection map~\eqref{eq_contraction}, which is defined as follows
\begin{equation}\label{eq_trans_constr_quad}
    \P_2(\x^{[2]}_{N_2}) = \T \, \P_{\text{TD}}(\x^{[2]}_{N_2}).
\end{equation}
In this context, the function $g_2(\x^{[2]}_{N_2})$ in~\eqref{subeq_switch} measures the vertical distance between the contact foot and ground, ensuring that touchdown occurs on the ground. The value function propagation~\eqref{eq_value_update_jump} across the model transition is then
\begin{subequations}\label{eq_value_update_modeltrans}
\begin{align}
    \s &= \boldsymbol{\Phi}_{\x} + \frac{\partial\P_{\text{TD}}}{\partial\x}^\top\T^\top\s',\\
    \S & = \boldsymbol{\Phi}_{\x\x} + \frac{\partial\P_{\text{TD}}}{\partial\x}^\top\T^\top\S' \T\frac{\partial\P_{\text{TD}}}{\partial\x},
\end{align}
\end{subequations}
where here the prime indicates the next step when the simple model starts. The equation~\eqref{eq_value_update_modeltrans} reveals that the trunk model is effectively used to set a terminal cost for the full model via a low-rank approximation, thus biasing the full model toward plans that are favorable for the long-term. %and reducing its dependence on a warm start.

We emphasize that the transition ~\eqref{eq_trans_constr_quad} is particular to the abstraction schedule at the time instant in Fig.~\ref{fig:hierarcy_legged}(a). Since the optimization window is shifting as MHPC proceeds, this transition constraint changes as well. For example, if we shift the overall planning horizon one gait mode to the future, the transition would occur at takeoff and would be described by $\P_2(\x^{[2]}_{N_2}) = \T\x^{[2]}_{N_2}$. Further, the abstraction schedule assigns two gait modes to each model, whereas in general it could be varied to acquire high performance. This prospect is explored further in Section~\ref{sec:Sim}.

A five-link biped robot is topologically equivalent to a planar quadruped (Fig.~\ref{fig:hierarcy_legged}) except that both legs are pinned below the trunk CoM. Consequently, the full model~\eqref{eq_fullmodel_smooth} and~\eqref{eq_fullmode_impact}, the simple model~\eqref{eq_trunk_model}, and the abstraction schedule for quadruped bounding are also applied to biped running.

\begin{figure}[t]
    \centering
    \includegraphics[width=\columnwidth]{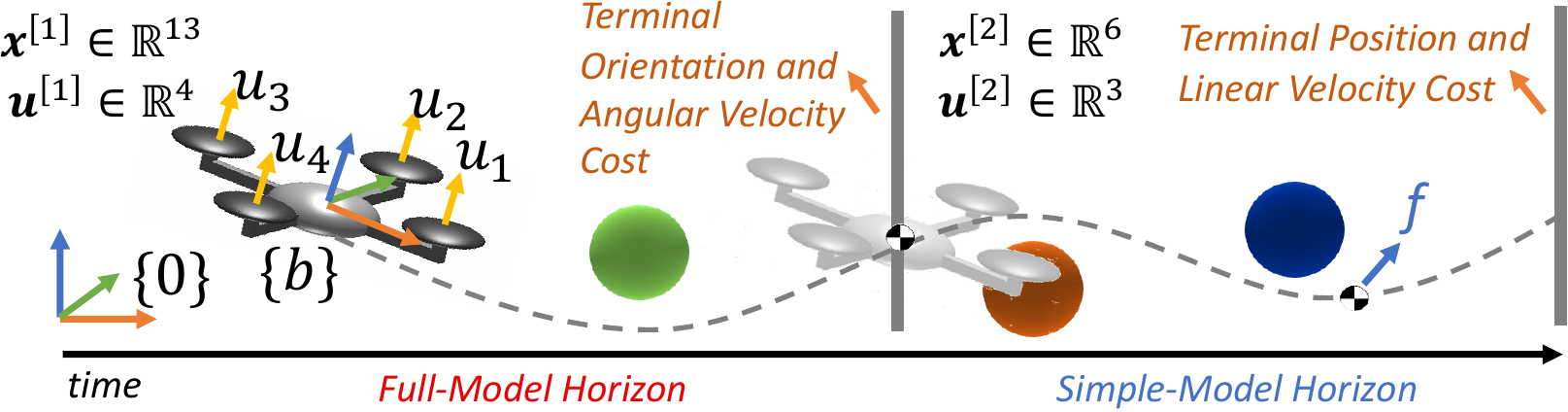}
    \caption{Quadrotor model hierarchy for motion planning around obstacles. The full model is followed by a point-mass model over the planning horizon.}
    \label{fig:Quadrotor}
    \vspace{-6px}
\end{figure}
\subsection{Quadrotor}
The last example explores the applicability of MHPC beyond legged robots. The quadrotor (Fig.~\ref{fig:Quadrotor}) is modeled as a rigid body with four thrust inputs, each at a moment arm $r$ from the CoM. The acceleration of the quadrotor CoM is modeled as
\begin{equation}
    {^b\a_{\CoM}} = {1\over m}\begin{bmatrix} 0 \\ 0 \\ u_1 + u_2 + u_3 + u_4 \end{bmatrix} + ^b\!\R_0 {}^0\! \g\end{equation}
with its rotational dynamics modelled as
\begin{equation}
    \Icom \,  ^b\dot{\bomega}  + ^b\bomega \times \Icom\, ^b\bomega  = \begin{bmatrix} r (u_2-u_4) \\ r (u_3 - u_1) \\ k_m (u_1+u_3-u_2-u_4) \end{bmatrix},
    \nonumber
\end{equation}
where the effort vector $\u = [u_1 , u_2 , u_3 , u_4]^\top$  contains the linear thrust of each rotor, $\Icom \in \mathbb{R}^{3\times3}$ is the Cartesian inertia of the drone about its CoM, $m$ its mass, and $^b\boldsymbol{\omega}$ its rotational velocity. In all cases, the pre-superscript indicates the frame that is used to express each vector. To approximate the yaw torque provided by the inertia of the counter rotating blades, $k_m$ is used as a ratio of the difference in thrust from the blades and the moment applied on the drone. %This approximates realistic behavior well. 
The full state of the drone is thus given by  $\x_f = [\quat, {}^b \bomega , {}^0 \p_{\CoM} , {}^b\v_{\CoM} ]^\top \in\Real ^{13}$ where $\quat$ gives its orientation quaternion, ${}^0 \p_{\CoM}$ is the position of the CoM, and ${}^b \v_{\CoM}$ its velocity. A fully actuated point mass model with linear dynamics is used as the simple model (Fig.~\ref{fig:Quadrotor}) with state $\x_s  =[{}^0 \p_{\CoM}^\top, {}^0 \v_{\CoM}^\top]^\top \in\Real ^{6}$. The input for the model is a single force vector in 3D.  %with the projection from full to simple model including a rotation in this case. 

To handle working with unit quaternions in the full model, a change of variables to minimal coordinates is employed where Cayley parameters are used to describe changes in orientation \cite{jackson2021planning}  at each iteration. 
%The Cayley parameters $\boldsymbol{\phi} \in \Real^3$ are related to an angle axis representation of rotation according to $\boldsymbol{\phi} = \hat{\mathbf{e}} \tan(\theta/2)$ where $\hat{\mathbf{e}} \in \Real^3$ specifies the axis and $\theta$ the angle. These parameters are also referred to as Rodrigues parameters or the Gibbs vector \cite{shuster1993survey} in other literature.  
%More specifically, the orientation at each timestep is represented as $\quat_k = \overline{\quat}_k  \otimes \boldsymbol{\varphi}( \boldsymbol{\phi}_k )$ where $ \overline{\quat}_k$ is the nominal quaternion at timestep $k$, $\otimes$ gives the quaternion product, and $\boldsymbol{\varphi}( \boldsymbol{\phi} )$ converts Cayley parameters to a quaternion as in \cite{Jackson20}.
%\[
%\boldsymbol{\varphi}( \boldsymbol{\phi} ) = %\frac{1}{\sqrt{1+\| \boldsymbol{\phi}\|^2}} %\begin{bmatrix} 1 \\ \boldsymbol{\phi} %\end{bmatrix}
%\]
Although Cayley parameters experience a singularity at a rotation of $\pi$ radians, since DDP makes small changes to the trajectory at each iteration, the singularity can be effectively avoided, even in cases when the nominal trajectory itself makes full rotations. The interested reader is referred to \cite{Jackson20} for detail.

\section{Simulation Results}
\label{sec:Sim}

In this section, we benchmark the performance of MHPC via simulation with the robots discussed in Section~\ref{sec:example_sys}. %The MHPC performance is evaluated in terms of disturbance rejection for the quadruped and biped, and in terms of cost reduction for the quadrotor. Computational cost is considered for various abstraction schedules with each system.

\subsection{Quadruped}\label{subsec_sim_quadruped}

We use a 2D model of the MIT Mini Cheetah \cite{katz2019mini} as the testbed. A bounding gait is simulated for four gait cycles, where the front-stance mode and the flight mode each runs for 72 ms, and the back-stance mode runs for 80 ms. This task is repeated for six MHPC abstraction schedules.

MHPC is configured to re-plan at every gait mode. We alert the reader that we count the planning horizon in gait modes in this section rather than in seconds as in Fig.~\ref{fig:MHPC_diagram} since each mode may have different timings. For simplicity, denote the number of gait modes in the overall planning horizon, full-model horizon, and simple-model horizon, as $n_o$, $n_f$, and $n_s$, respectively. We fix $n_o = 8$ for five abstraction schedules where $n_f$ is considered at values in $\{0,2,4,6,8\}$, and $n_s = n_o - n_f$. A sixth schedule uses $n_o = n_f = 4$ with no simple-model horizon. In the rest of this paper, MHPC($n_f$, $n_s$) denotes the abstraction schedule ($n_f$, $n_s$). When $n_f \neq 0$, MHPC generates joint toques that are directly applied to the robot. When $n_f = 0$, MHPC degenerates to simple-model MPC and generates the GRF as shown in Fig.~\ref{fig:hierarcy_legged}(a). In the latter case, we use $\taub = \J^\top\mathbf{F}$ to get joint torques for the stance leg, and a swing foot controller as in \cite{di2018dynamic} for the swing leg. A heuristic controller used by \cite{li2020hybrid} is employed to warm start the full-model plan, and the simple-model plan is initialized with zeros. The time steps for dynamic simulation and MHPC are both 1 ms. For each optimization, both the outer loop and inner loop of HSDDP are terminated at a maximum of 3 iterations regardless of convergence.

\begin{figure}[t]
    \ifcaptionmod
    \captionsetup[subfigure]{font=scriptsize,labelfont=scriptsize}
    \fi
    \centering
    \includegraphics[width = 0.7 \columnwidth]{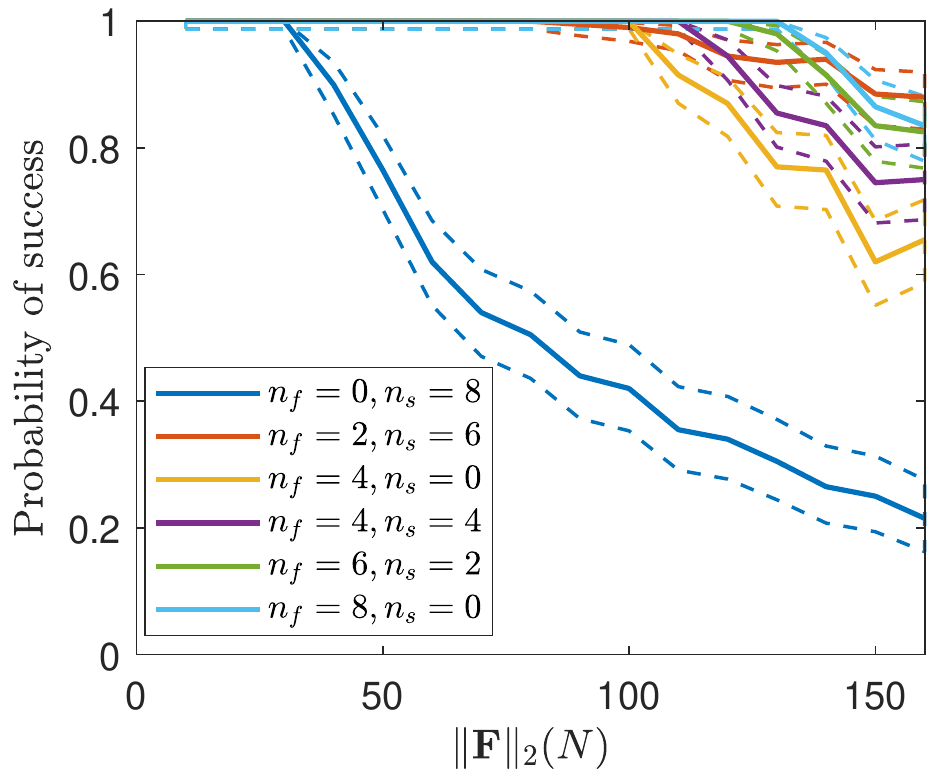}
    \caption{Robustness evaluation of MHPC for six abstraction schedules as applied to the quadruped robot.}
    \label{fig:robustness_quadruped}
    \vspace{-5px}
    \end{figure}

\begin{table}[t]
    \centering
    \caption{Normalized computation times of each MHPC configuration for the quadruped bounding and biped running examples. The first row represents the abstraction schedule $(n_f, n_s)$.}    
    \label{tab:comp_times}
    \begin{tabular}{*{8}{|c}|}
    \hline
    \multicolumn{2}{|c|}{}&(0,8) &(2,6) &(4,0) &(4,4) &(6,2) &(8,0)  \\
    \hline
    \multirow{2}{*}{Quad} & Avg & 0.100 & 0.530 & 0.623 & 0.678 & 0.855 & 1\\
    \cline{2-8}
    & Std & 0.016 & 0.027 & 0.030 & 0.032 & 0.040 & 0\\
    \hline
    \multirow{2}{*}{Biped} & Avg & $\times$ & 0.400  &0.621 &0.530  & 0.770 & 1\\
    \cline{2-8}
    & Std & $\times$ & 0.070 & 0.042 & 0.085  & 0.008 & 0\\
    \hline
    \end{tabular}
\end{table}

    \begin{figure}[t]
    \centering
    \includegraphics[width = 0.7 \columnwidth]{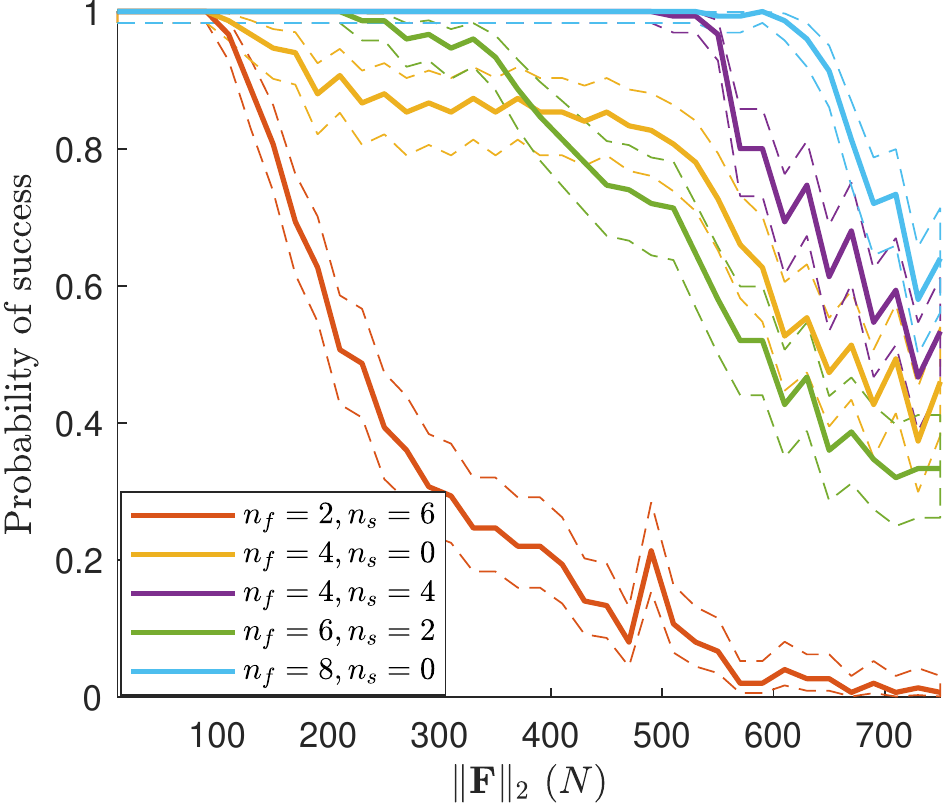}
    \caption{Robustness evaluation of MHPC for five abstraction schedules as applied to the biped robot.}
    \label{fig:robustness_biped}
    \vspace{-15px}
    \end{figure}

Figure~\ref{fig:robustness_quadruped} statistically quantifies the robustness of MHPC for each abstraction schedule, evaluated by the likelihood of rejecting push disturbances. Disturbances are applied on the trunk for 30 ms in the second flight, with magnitude varying between $10- 160 N$, and location and direction randomly sampled from a uniform distribution. The probability of success is estimated based on 200 random tests. The solid curves represent the estimated probability of success, around which the dashed curves indicate a $95\%$ confidence interval. The top right corner indicates a higher possibility of rejecting larger disturbances, whereas the left bottom corner indicates more likely failures for small disturbances. The configuration MHPC(0,8) (simple-model MPC) has the worst performance, since it starts to fail at disturbances of 30 $N$. The configuration MHPC(8,0) (whole-body MPC) has the best robustness, since it can reject disturbances as much as 130 $N$. The robustness of other MHPC configurations are above that of MHPC(0,8), demonstrating that incorporating full-model planning would increase the robustness. The result of MHPC(2,6) shows that even assigning a short interval of the time horizon to the full model could significantly improve the controller's robustness. Comparison between MHPC(4,4) and MHPC(4,0) reveals that the performance is improved by extending the horizon with the simple model. In this sense, both adding whole-body planning to a simple-model scheme, and adding simple-model planning to a whole-body scheme offer performance benefits. 

Table~\ref{tab:comp_times} summarizes the average and standard deviations of the normalized computation times for each MHPC configuration, obtained assuming no disturbances. Since MHPC re-plans at every gait mode, sixteen optimizations are averaged for each MHPC configuration. The average computation times are normalized by that  of MHPC(8,0). This normalization is done as the implementation of HSDDP is in MATLAB, so the absolute time required is less meaningful. The speed-ups reported would be expected to translate to other more efficient implementations. Unsurprisingly, MHPC(0,8) (simple-model MPC) has the lowest computational cost, whereas MHPC(8,0) has the highest, and the computation time of other schemes increases with $n_f$. This is because the full model has the highest state dimension, and the computational complexity of DDP iterations are cubic in state dimension. Note that MHPC(2,6) achieves performance as good as whole-body MPC, while its computation efficiency is the second best. We conclude that with the proper schedule, MHPC could achieve high performance rivaling whole-body MPC, while significantly lowering computational cost.

\subsection{Biped}

The biped testbed used in this work is the simulated five-link planar biped robot Ernie \cite{fevre2019terrain} at the Univ.~of Notre Dame. MHPC is applied for four gait cycles with a target running speed of 1.5 $m/s$. The stance modes and the flight modes run for 110~ms and 80~ms, respectively. The outer loop and inner loop of DDP are terminated at maximum of 3 and 15 iterations, respectively. All other simulation parameters (e.g., simulation and MHPC time steps, and overall planning horizon) are identical to the quadruped simulation in Section~\ref{subsec_sim_quadruped}.

Figure~\ref{fig:robustness_biped} depicts the probability of success of MHPC for five abstraction schedules in response to disturbances 0-750~$N$. The MHPC(0,8) (simple-model MPC) is not shown here since the authors were unable to obtain a stable running gait for the biped with this approach. This finding is attributed to the fact that the balance problem for a biped is typically more difficult than that for a quadruped, and more whole-body details need to be considered in designing a stabilizing controller. It is observed that MHPC(4,4) and MHPC(8,0) fully reject disturbances under 500 $N$ and 600 $N$, respectively, with which the resulting velocity disturbances are 0.5 $m/s$ and 0.6 $m/s$. Further, Table~\ref{tab:comp_times} shows that the computational cost of MHPC(4,4) is roughly half that of MHPC(8,0), demonstrating that by selecting a proper abstraction schedule, MHPC could achieve nearly comparable disturbance rejection performance to whole-body MPC with significantly lower computational cost. We note that MHPC(4,4) takes less computation time than MHPC(4,0), because MHPC(4,4) takes less DDP iterations to converge. 

\subsection{Quadrotor}

\begin{figure}[b]
    \centering
    \includegraphics[width=.65\columnwidth]{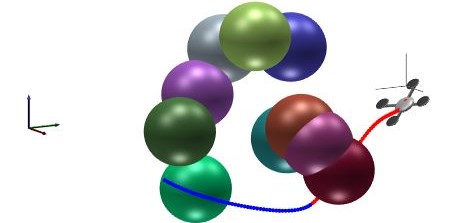}
    \caption{Quadrotor tests planned point-to-point motions from a starting configuration (right) to a goal configuration (left). The red and blue lines show the full-model and simple-model horizons, respectively.}
    \label{fig:Quadrotor_Animation}
\end{figure}

\begin{figure}[t]
    \centering
    \includegraphics[width=.95\columnwidth]{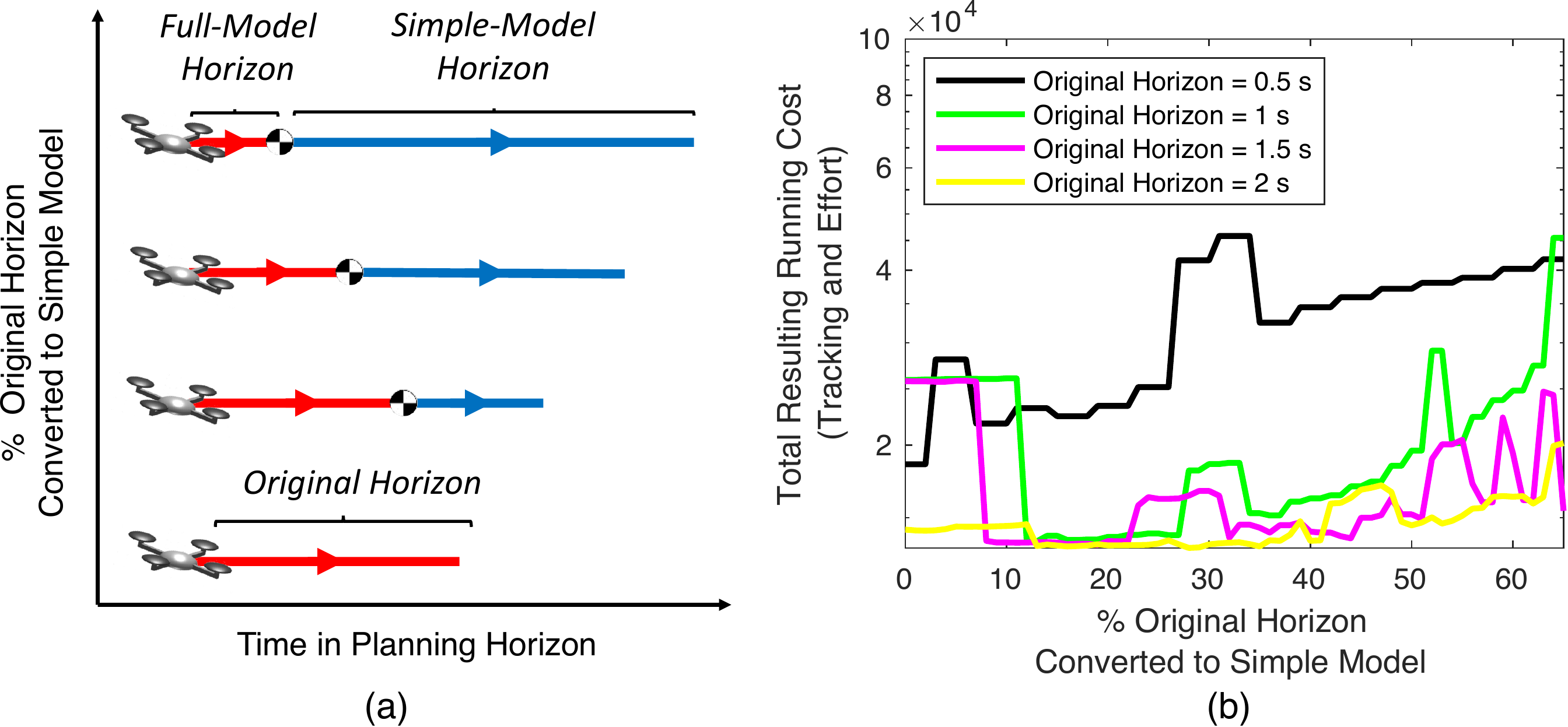}
    \caption{(a) Different MHPC planning horizons with equivalent computational requirements (b) Resulting performance as MHPC horizon composition is varied as depicted in Fig.~\ref{fig:Quadrotor_Results}(a).}
    \label{fig:Quadrotor_Results}
    \vspace{-10px}
\end{figure}

 To benchmark MHPC on the quadrotor, a point-to-point trajectory avoiding obstacles is planned and executed (Fig.~\ref{fig:Quadrotor_Animation}). The spherical obstacles are randomly generated then fixed across all trials. MHPC is solved in a receding horizon fashion and the entire simulation lasts 5~$s$ with a 0.02~$s$ time step. The initial guesses are all zeros for both models. The running costs assigned to the full and simple models are quadratic. The full model includes a quaternion orientation term $(1- (\quat_d^\top \quat)^2)$, minimized at either the desired quaternion $\quat_d$ or its antipode $-\quat_d$, which provides the same orientation. A relaxed barrier method is used to impose non-collision constraints with the obstacles considering a spherical volume approximation for the quadrotor. A quadratic terminal cost is assigned to the rotational error at the end of the full-model trajectory, and the final terminal cost is the infinite-horizon LQR cost for the simple model, ignoring obstacles. To evaluate the benefit of MHPC, we first implement MPC with the full model only, and then evaluate the performance when a portion of the horizon is instead allocated to the simple model (Fig.~\ref{fig:Quadrotor_Results}(a)). The length of horizon assigned to the simple model is adjusted so that one iteration of DDP takes the same amount of time in all cases. With this strategy, a horizon reduction of one time step for the full model conservatively gives 2.5 additional time steps for the simple model. Thus, allocating a portion of the horizon to the simple model enables longer-horizon planning overall. At each re-planning step, DDP is warm-started using the previous plan and runs for a maximum of five iterations to simulate real-time behavior.

The total running costs for each simulation are given against horizon composition in Fig.~\ref{fig:Quadrotor_Results}(b).
The total cost reaches a minimum as the original full-model planning horizon is replaced by a longer simple-model horizon, except for the 0.5 $s$ horizon which under-performed. Interestingly, MHPC results in a similar optimal performance between the original horizons of $1\sim 2$ $s$. This result shows that even for computationally cheaper setups (e.g., $1$ $s$ original horizon), MHPC can exceed the performance of more computationally costly full-model MPC instances (e.g., $2$ $s$ original horizon). Overall, these results show that for a fixed computation time budget, MHPC enables longer-horizon prediction with reliably improved performance over full-model MPC. The companion video further illustrates these effects.

\section{Experimental Validation}
\label{sec:Exp}

In this section, the developed MHPC controller is implemented on the Mini Cheetah hardware \cite{katz2019mini} to achieve bounding with a desired forward speed of 1.5 $m/s$ while avoiding a gap. The gap has width $0.4$ $m$, and is located at 1.0~$m$ relative to the origin of the world frame. The control and estimation loops run at 1000~Hz on hardware. The MHPC handles the gap using an inequality constraint that enforces the feet to be above a sixth-order rational function that is illustrated as grey solid line in Fig.~\ref{fig:MHPC_illusrative}. Since leg details are ignored in the simple-model plan, virtual legs are fixed to the trunk to enforce the gap constraint.

\begin{figure}[!t]
    \centering
    \includegraphics[width = 0.9\linewidth]{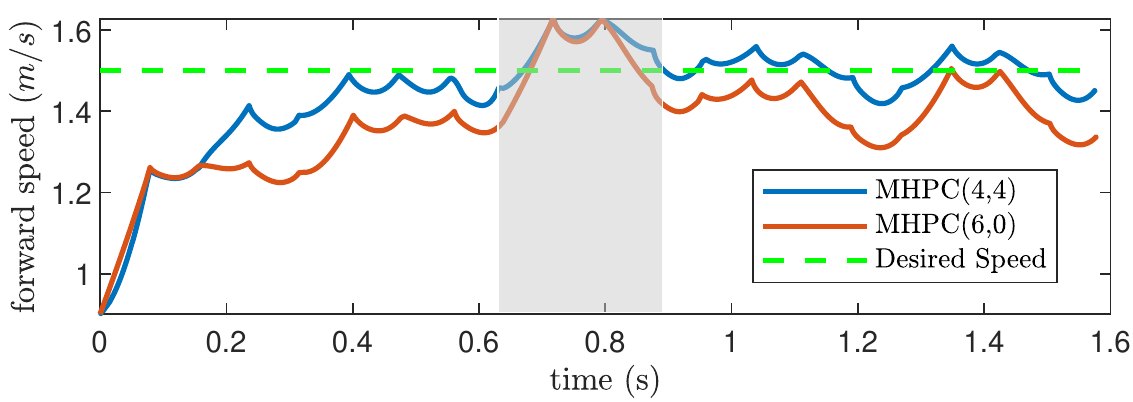}
    \caption{Forward speed for Mini Cheetah jumping over the gap. The blue solid line shows the result for MHPC(4,4) and red solid line shows the result for MHPC(6,0). The dashed line represents the desired forward speed 1.5 $m/s$ and the shaded area indicates the time period when jumping takes place.}
    \label{fig:forward_speed}
\end{figure}

\begin{figure}[t]
    \centering
    \includegraphics[width = 0.9\linewidth]{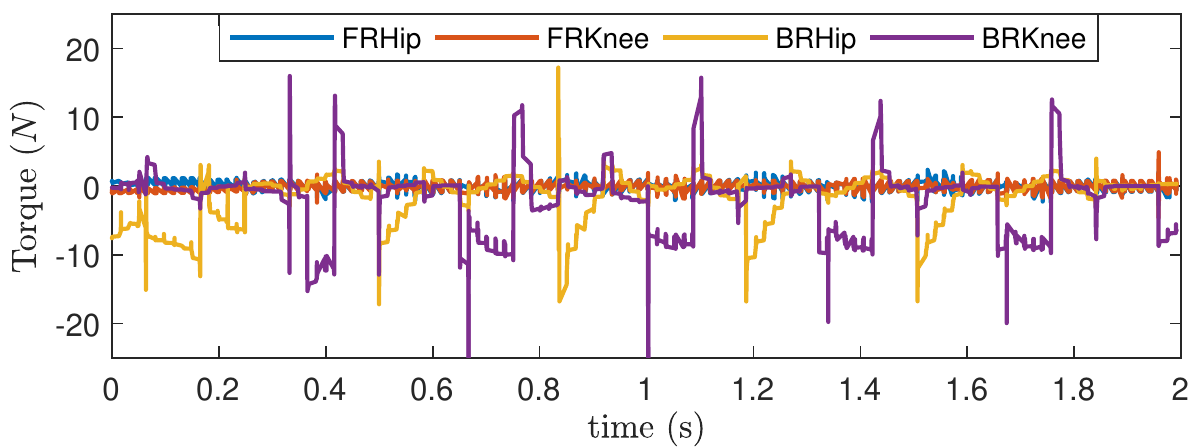}
    \caption{Hip and knee torques for the front right and back right legs, obtained in a dynamics simulator designed for the MIT Mini Cheetah.}
    \vspace{-10px}
    \label{fig:torque}
\end{figure}

Two configurations with similar computation demands, MHPC(4,4) and MHPC(6,0), are first evaluated in simulation. 
%and the policy from MHPC(4,4) is implemented on the hardware. 
We alert the readers that the maximum inner and outer loop iterations of DDP are set to seven, as opposed to three used in normal bounding, since satisfying the gap constraint requires more iterations. 
Fig.~\ref{fig:forward_speed} depicts the forward speed of the robot for both cases with the shaded area indicating the gap jump. The figure shows that the robot accelerates to a higher speed before the jump in both cases, however, the MHPC(4,4) has smaller tracking error than MHPC(6,0), and its longer horizon enables it to begin accelerating earlier. Further, the total roll-out costs are $1.9\times 10^4$ and $2.6\times 10^4$, respectively, for MHPC(4,4) and MHPC(6,0), demonstrating that MHPC(4,4) has better performance, while the averaged computation times of both are approximately the same.

A control policy computed from MHPC(4,4) offline is then implemented on the MIT Mini Cheetah hardware online. The feed-forward part of the computed control policy is used directly for the joint actuation, whereas the feedback gain in \eqref{eq_optdu} associated with the trunk state is scaled by 0.1 due to the inaccurate estimation of the trunk state obtained from the current state estimator. In addition to DDP gains, a PD-controller $\taub = -k_p(\qs_{\text{joint}} - \qs_{\text{joint,n}}) - k_d(\Dot{\qs}_{\text{joint}} - \Dot{\qs}_{\text{joint,n}})$ is added to 
robustify joint tracking where $k_p = 3 (N\cdot m)/rad$ and $k_d = 1 (N\cdot m)/rad/s$. Further, since the controller is based on a planar quadruped, a roll and a yaw controller as in \cite{park2017high} are employed to regulate roll and yaw errors. 

\begin{figure*}[!h]
 \centering
    \includegraphics[width = 0.91\textwidth]{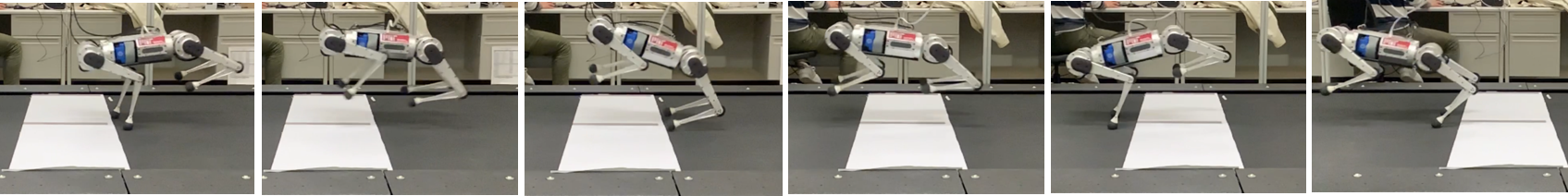}
    \caption{Snapshots of the Mini Cheetah executing a bounding motion while avoiding a mock gap (white foam board).}
    \label{fig:bounding_exp}
    \vspace{-12px}
\end{figure*}

Figure~\ref{fig:torque} depicts the hip and knee torques for the front right and back right legs. The data are obtained in a dynamics simulator that is designed to mimic the real-world experiment. Figure~\ref{fig:bounding_exp} shows time-series snapshots of the hardware implementation. The white foam of width 0.4 m is meant to mimic the gap, and is located at 1.0 m along the forward direction in world frame. Once the front-stance phase ends, the robot quickly swings its front legs to avoid the board and prepares itself for the jump. The robot was manually pulled up after crossing the gap when reaching the end of the treadmill. The accompanying video shows another experiment where a normal bounding gait is executed. The robot was pulled up after ten gait cycles for the same reason. The running surface remains static since it will otherwise create a yaw torque on the robot at touchdown while accelerating, and the current 2D plan is stabilized by a rudimentary yaw controller. It is anticipated that online replanning (and potentially planning in 3D) will address this issue in future experiments.

\section{Conclusions and Future Works}
\label{sec:Conclusion}
This work presented a new Model Hierarchy Predictive Control (MHPC) architecture that combines the benefits of full-model MPC and simple-model MPC. The simple model is used in long-term planning to set a low-rank approximation of the optimal cost-to-go for the terminal cost of the full model. This approach automatically biases the full-model optimization toward a favorable long-term plan as opposed to heuristic terminal costs authored into whole-body MPC. Further, the simple model has a low dimension and can reduce the overall computational cost. Employing a full model in near-term planning, even for a short period, is shown to significantly improve push robustness compared to MPC with a simple-model alone. Simulation results have shown that with the proper abstraction schedule, MHPC achieves high disturbance rejection comparable to whole-body MPC while maintaining significantly lower computational requirements. Converting a portion of a whole-body MPC horizon to a longer template horizon has also shown to extend predictions, improving MPC outcomes in challenging environments.

Currently, the MHPC implementation is computed offline and executed online. Future work will focus on online computation. Our preliminary work in this direction aims for a second-order algorithm that efficiently computes needed partials by exploiting structure in the equations of motion. %Future work will also address the yaw control issue for the planar quadruped and unlock more dynamic behaviors by employing a 3D model.

\ifarxiv
\bibliography{paper.bbl}
\else
\bibliographystyle{IEEEtran}
\bibliography{paperbib}
\fi

\end{document}